\documentclass[conference,letterpaper]{IEEEtran}
\IEEEoverridecommandlockouts
\usepackage[dvips]{graphicx}  
\usepackage{amssymb,color,theorem}
\usepackage{subfig}
\usepackage{algorithm}
\usepackage{algpseudocode}
\usepackage[fleqn]{amsmath}
\usepackage{booktabs}  
\usepackage{threeparttable}  
\usepackage[ruled, algo2e]{algorithm2e}                                                
\usepackage{multirow}\usepackage[left=0.71in,top=0.94in,right=0.71in,bottom=1.18in]{geometry}
\usepackage{mars}

\usepackage[obeyspaces]{url}

\IEEEoverridecommandlockouts                              

\title{\huge Furniture Free Mapping using 3D Lidars}

\author{\authorblockN{Zhenpeng He, Jiawei Hou and S\"{o}ren Schwertfeger}
	\authorblockA{\textit{School of Information Science and Technology}\\
		\textit{ShanghaiTech University}\\
		\textit{Shanghai, China}\\
		\textit{\{hezhp, houjw, soerensch\}@shanghaitech.edu.cn}\\}%
}%

\begin{document}
\marsPublishedIn{Accepted for: ROBIO2019} 		

\marsVenue{International Conference on Robotics and Biomimetics (ROBIO) 2019}

\marsYear{2019}

\marsPlainAutors{Zhenpeng He, Jiawei Hou and S\"oren Schwertfeger}


\marsMakeCitation{Furniture Free Mapping using 3D Lidars}{IEEE Press}

\marsDOI{\url{}}

\marsIEEE{}


\makeMARStitle

\maketitle
\begin{abstract}
	
	Mobile robots depend on maps for localization, planning, and other applications. In indoor scenarios, there is often lots of clutter present, such as chairs, tables, other furniture, or plants. While mapping this clutter is important for certain applications, for example navigation, maps that represent just the immobile parts of the environment, i.e. walls, are needed for other applications, like room segmentation or long-term localization. In literature, approaches can be found that use a complete point cloud to remove the furniture in the room and generate a furniture free map. In contrast, we propose a Simultaneous Localization And Mapping (SLAM)-based mobile laser scanning solution. The robot uses an orthogonal pair of Lidars. The horizontal scanner aims to estimate the robot position, whereas the vertical scanner generates the furniture free map. There are three steps in our method: point cloud rearrangement, wall plane detection and semantic labeling. In the experiment, we evaluate the efficiency of removing furniture in a typical indoor environment. We get $99.60\%$ precision in keeping the wall in the 3D result, which shows that our algorithm can remove most of the furniture in the environment. Furthermore, we introduce the application of 2D furniture free mapping for room segmentation. 
	
\end{abstract}

\begin{keywords}
	Furniture Free Map, SLAM, Semantic, Real-time.
\end{keywords}

\section{Introduction}
\label{sec::introduction}

Indoor maps of buildings are critical in many robotics application such as navigation \cite{taneja2016algorithms}, localization \cite{ryu2016grid}, planning \cite{pinheiro2015cleaning}, room segmentation \cite{bormann2016room, hou2019area} or topological representations \cite{Schwertfeger2015Map}. One difficult challenge in room segmentation is the over-segmentation caused by clutter in the generated map. Some topological maps are created based on segmented maps, which can be used to reduce the computational complexity of large scenarios. In indoor environments, objects like furniture frequently occlude the wall surfaces, which puts more challenges on segmentation. There are some methods to deal with cluttered and occluded environments. Adan et al. \cite{adan20113d} used a histogram of the point cloud to detect the wall and generate the furniture free map. Babacan et al. \cite{babacan2017semantic} introduced a Convolutional Neural Network (CNN) to filter out the noise in the environment. Mura et al. \cite{mura2014automatic} proposed an efficient occlusion-aware process to extract planar patches as candidate walls, separating them from clutter and coping with missing data. These static solutions need a complete point cloud. Nevertheless, real-time strategies are increasingly required in this field. 

\begin{figure}[htbp!]
	\centering
	\subfloat[One room point cloud with lots of work cubicles.]{
		\label{fig:res_room_origin}
		\includegraphics[height=3.4cm]{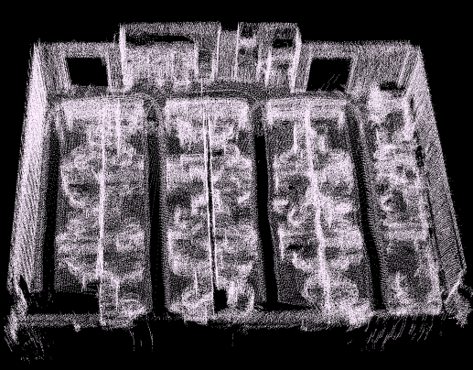}}
	\subfloat[Point cloud of the same room with only walls (green) and doors (blue).]{
		\label{fig:res_room_result}
		\includegraphics[height=3.4cm]{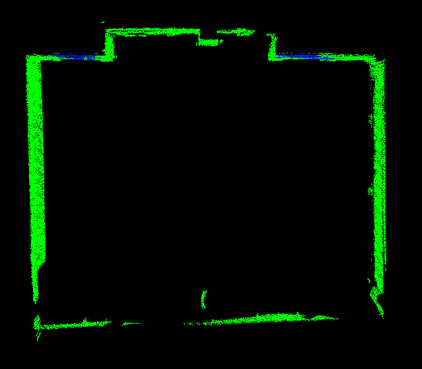}}
	\caption{Furniture free mapping example.}
	\label{fig:noise_remove}
\end{figure}

\begin{figure*}[t]
	\centering
	\includegraphics[width=0.90\linewidth]{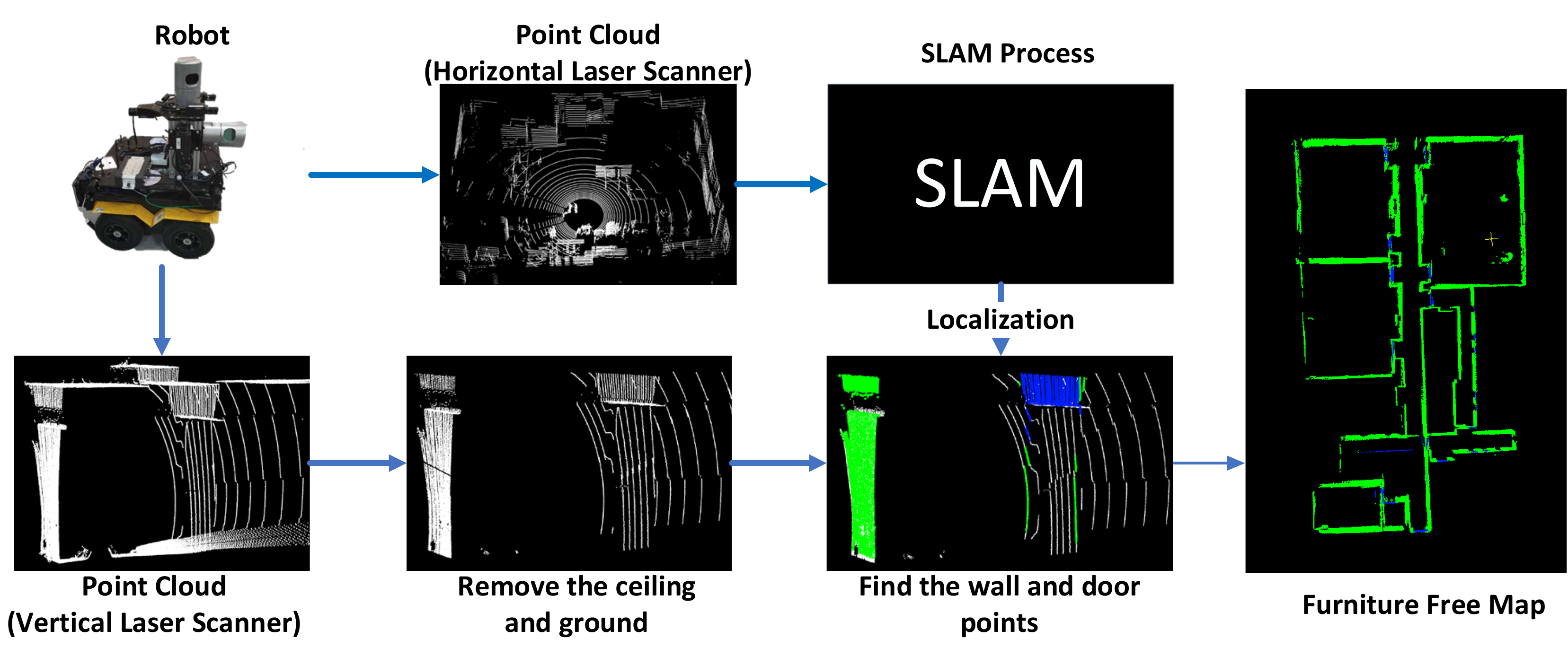}
	\caption{Pipeline of the proposed algorithm.}
	\label{fig:pipeline}
\end{figure*}

In this study, we propose a real-time method for a mobile robot which has two 3D Lidar (LIght Detection And Ranging) sensors: a horizontally scanning and a vertically scanning laser sensor. The robot additionally has nine color cameras, an IMU, and odometry. All sensors are fully synchronized \cite{Chen2019Towards} and calibrated \cite{Chen2019Heterogeneous}. The horizontal scanner is used for Simultaneous Localization And Mapping (SLAM), from which we utilize only the pose information of the robot, while the vertical sensor is key to 3D mapping and our semantic labeling algorithm. There are three steps in our approach, which are all applied to the vertical scanner: 1) point cloud rearrangement, 2) wall plane detection, and 3) semantic label reasoning. In the first step, we remove the ceiling and floor of each frame of the point cloud. Then we detect the wall candidate elements and use a growing algorithm to compute the wall plane coefficient. In the final step, a set of reasoning chains is used to label each point as the wall, ceiling, and others. According to the semantic labels, we can remove the movable objects and only keep the wall points in the result. It is important to mention that an explicit detection of openings (doors) is needed to create 2D maps that correctly represent the connections (doors) between the environments. The pipeline process for our proposed method is shown in Fig. \ref{fig:pipeline}.

Our wall plane detection method is based on the assumption that walls are planar and perpendicular to the ground. Artificial indoor scenes often exhibit repeated regularity, especially in the vertical direction. So we also believe that identifying semantic labels in the indoor building is mainly a detection problem, rather than a common segmentation problem.

The outputs of our method are a 3D semantic point cloud and a 2D furniture free grid map. The experiment shows that our method has a high precision for finding the wall plane. We also get good performance in noise removal, which is shown in Fig. \ref{fig:noise_remove}. 
The first experiment shows the semantic classification accuracy of the 3D point cloud. The second experiment compares the results of the furniture free mapping with a hand-made 2D clean map. The third experiment demonstrates our approach in a 2D room segmentation application.  

The main contributions of this paper are as follows:
\begin{itemize}
	\item An algorithm for large-scale indoor point cloud classification.
	\item Derived from that a method to generate furniture free 2D grid maps. 
	\item Experimental analysis of the algorithm and application to room segmentation.
\end{itemize}

\section{Related Work}
\label{sec::related}

\subsection{Indoor Modeling and Reconstruction}

Furniture free mapping is quite related to indoor modeling and reconstruction, because it basically generates the floor plan of a building. Methods for creating such models can be divided into two types: those that use one complete point cloud of the environment as the input and those that work with individual laser scans, typically in real time. Those models are usually higher-order representations of the environment, like planes, meshes or lines. 

Real time mapping is most often embedded in SLAM. Chen et al. \cite{chen2016real} are using a vertical laser scanner to build the model with a SLAM process. \cite{pintore2014effective} applied a mobile phone and a set of fixed actions to produce a 2D floor plan and a representative 3D model. Kim et al. \cite{kim2012interactive} used the hand-held depth camera and allowed the user to freely move through a building to capture its important architectural elements. This algorithm needs to initialized by pointing the sensor to a corner. Turner et al. \cite{turner2014floor} triangulated sampling points on a 2D planar map and separated these triangles into the interior and exterior sets. The exterior sets are outside of the building, so only the interior sets are kept to create the model.

Creating models from big, static point clouds is well researched. The approach in \cite{sanchez2012planar} utilized random sample consensus (RANSAC) and plane normal orientations to extract structures of the building, such as floor, ceiling, and walls. The study of Babacan \cite{babacan2016towards} used minimum description length (MDL) hypothesis ranking to extract the floor blueprint. Oesau  \cite{oesau2014indoor} presented an energy minimization function for different segment rooms in 2d maps. Okorn \cite{okorn2010toward} used a point histogram and an iterative Hough transform algorithm to generate the map.

\section{Method}
\label{sec::method}

In our method, the goal is to distinguish the points of the wall, doors, and furniture. These points belong to different regular and irregular components in the point cloud. The regular components in the indoor point cloud can be divided into two geometrically significant clusters, according to the distribution of surface normals: horizontal structure and vertical structure. The horizontal structure includes mainly the ceiling and the floor. As mentioned above, we are using the vertical laser scanner of the robot for this part, so typically, there will always be a ceiling and floor element to be found. The vertical structure consists primarily of walls. Irregular components have no apparent geometric distribution. Our pipeline to classify these mixed data is to start by extracting the ceiling and floor from the horizontal structure, and further determine wall planes from the remaining cloud. Semantic classification is then conducted according to the relative pose between the object and the wall, as shown in  Fig. \ref{fig:pipeline}.

\subsection{Point Cloud Rearrangement}
\label{sec::rearrangement}

\begin{figure}[tb]
	\centering
	\includegraphics[width=0.85\linewidth]{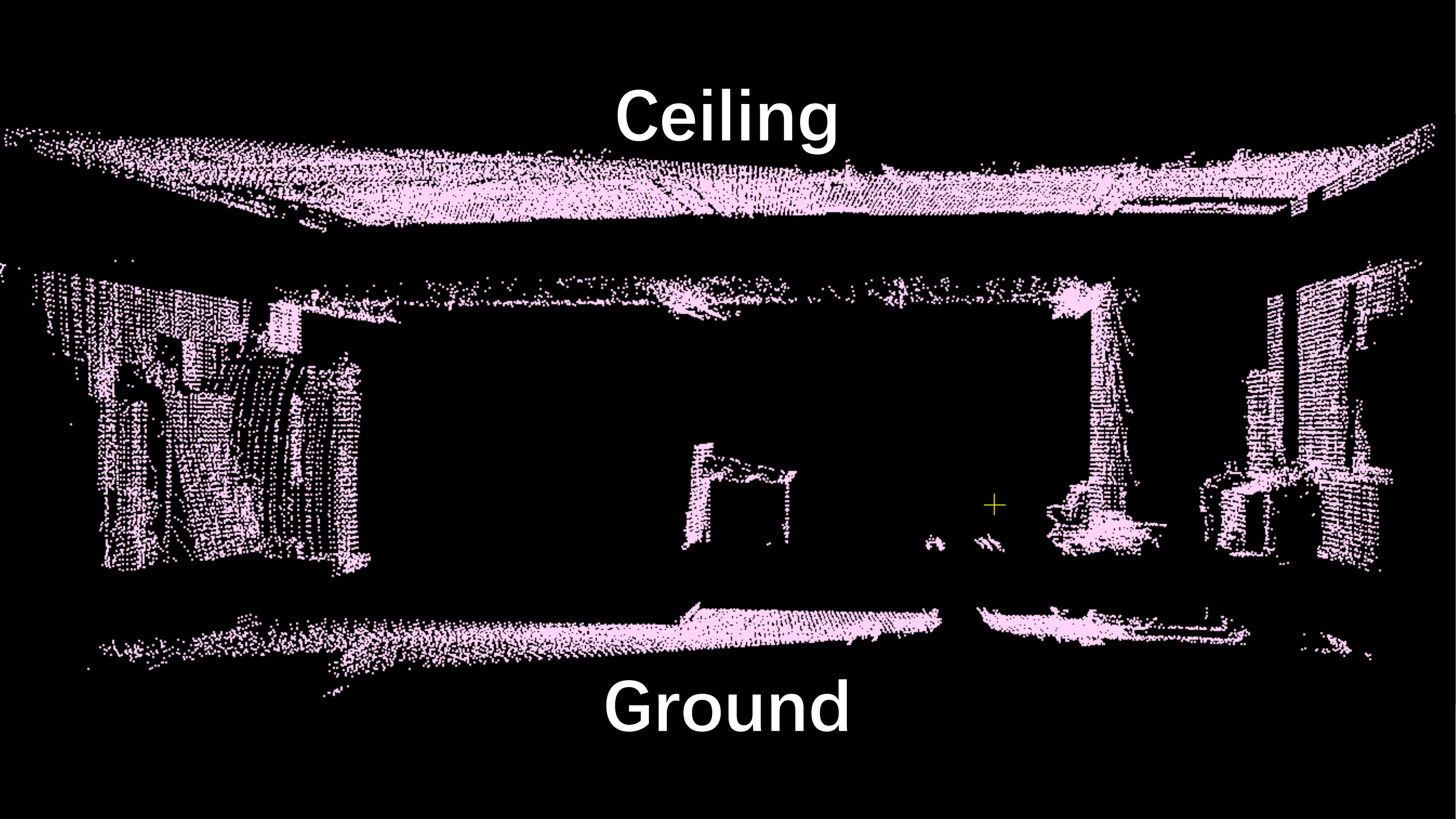}
	\caption{Point cloud from the vertical laser scanner after removing the ceiling and floor.}
	\label{fig:pointcloud_seg}
\end{figure}

\begin{figure}[bt!]
	\centering
	\subfloat[Beam layout of the laser scanner.]{
		\label{fig:mode1}
		\includegraphics[width=0.5\linewidth ,height=2.7cm]{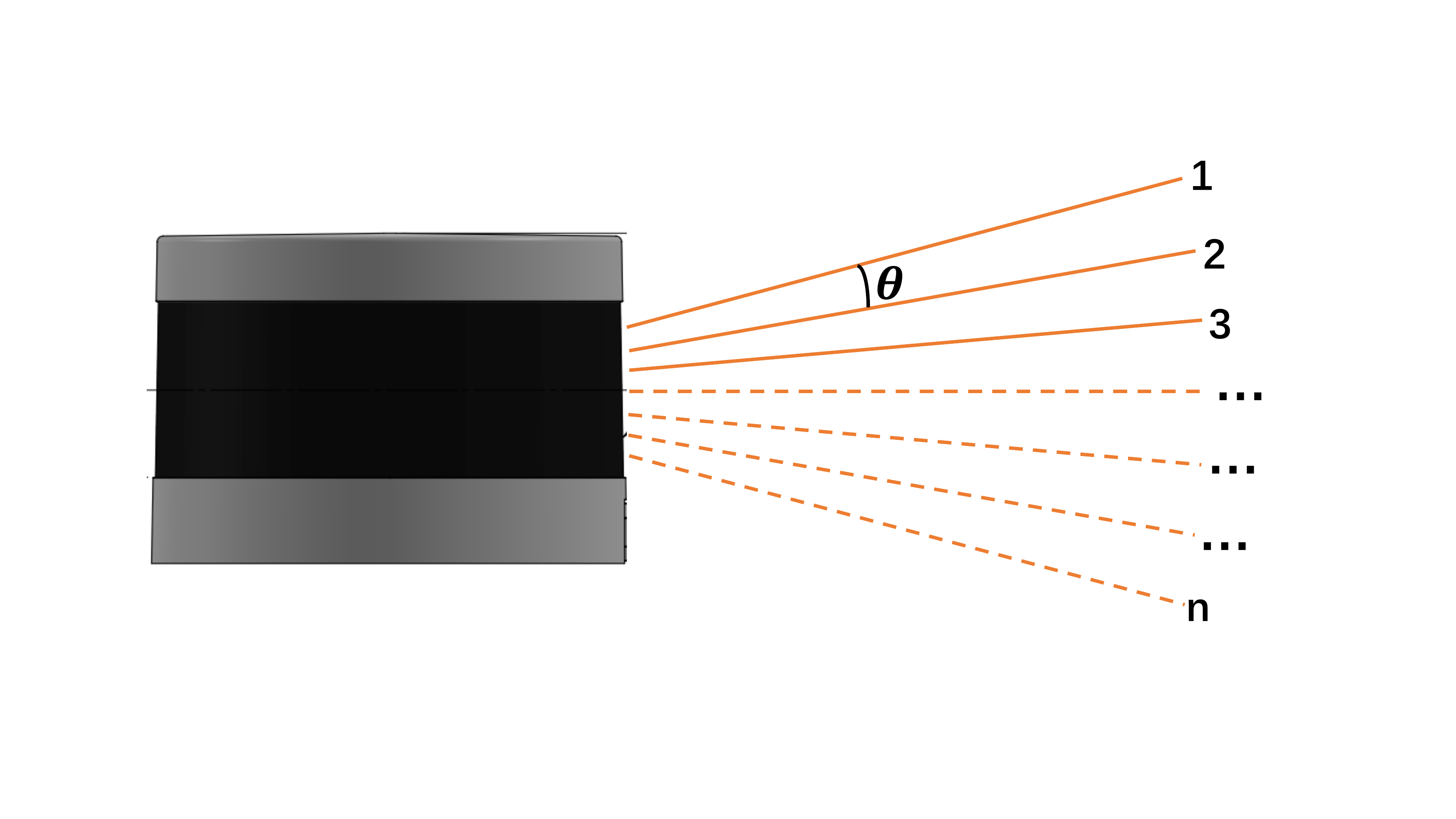}}
	\subfloat[Return mode.]{
		\label{fig:mode2}
		\includegraphics[width=0.5\linewidth ,height=2.7cm]{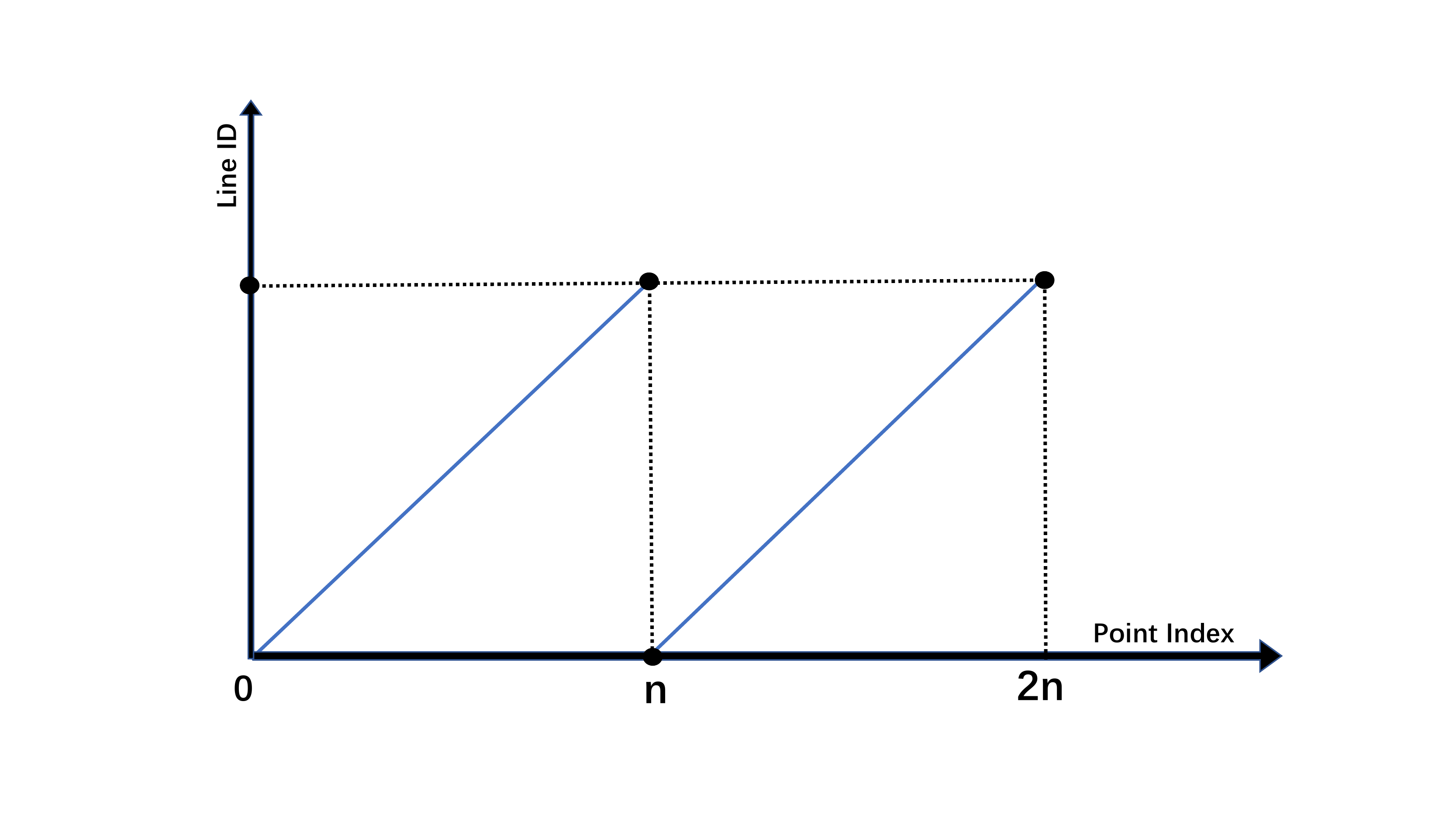}}
	\caption{Beams of the 3D laser sensor. }
\end{figure}

Firstly, we want to classify the position of the floor and ceiling. On the one hand, our algorithm is running on a ground robot so that we can straightforward locate where the floor is. We use a height filter to remove the floor from each frame of the scanner. We also assume that the ceiling is parallel to the floor. We use a Random Sample Consensus (RANSAC) algorithm to extract the biggest plane whose surface normal is orthogonal to the floor. As shown in Fig. \ref{fig:pointcloud_seg}, the remaining point cloud is mainly consisting of vertical structures after removing the ceiling and floor. We divide the point cloud in the vertical direction to many lines, as shown in Fig. \ref{fig:point_line}. The 3D laser scanner is composed of discrete laser emitters (32 for our robot). We extract the points from each emitter and then get an ordered point cloud. Fig. \ref{fig:mode1} shows the layout of the beams of the laser scanner. There is a fixed angular resolution in the point cloud. We sort these points based on their angle value $a$ to the position of the laser scanner. Let $\mathcal{P} = \{p_i, i \in \mathcal{N}\}$ be the set of points in a point cloud, which is from one frame of the vertical laser scanner. $(x_i, y_i, z_i) $ are the measured coordinates of $p_i$. Then the angle value $a$ is computed by:
\begin{equation}
a = \arctan{\frac{\sqrt{x_i^2 + z_i^2}}{y_i}}
\end{equation}
Points that have the same angle $a$ are categorized into the same line. There are $n$ lines of the laser scanner, such that we can divide each frame of point cloud to $2n$ vertical point lines in Fig \ref{fig:mode2}. To speed up the further computations we then, based on the angle, subsample the point lines to have exactly 200 points.

\subsection{Wall Plane Detection}

In this step, we extract wall planes from point lines $\mathcal{L} = \{l_i, i \in \mathcal{N}\}$. Firstly, we find the wall candidate elements in those point lines. There are both noise and vertical structures in a line. We compute the forward difference of adjacent points along the line to determine vertical structures. Let $p_j ,p_{j-1} \in l_i$ be the neighboring points. The forward difference $d$ is computed by:
\begin{equation}
d = \bigg| \frac{\sqrt{x_j^2 + y_j^2} - \sqrt{x_{j-1}^2 + y_{j-1}^2}}{z_j - z_{j-1}} \bigg|
\end{equation}

\begin{figure}[tb]
	\centering
	\subfloat[Extraction of a point line from the point cloud of one scan.]{
		\label{fig:point_line}
		\includegraphics[width=0.65\linewidth]{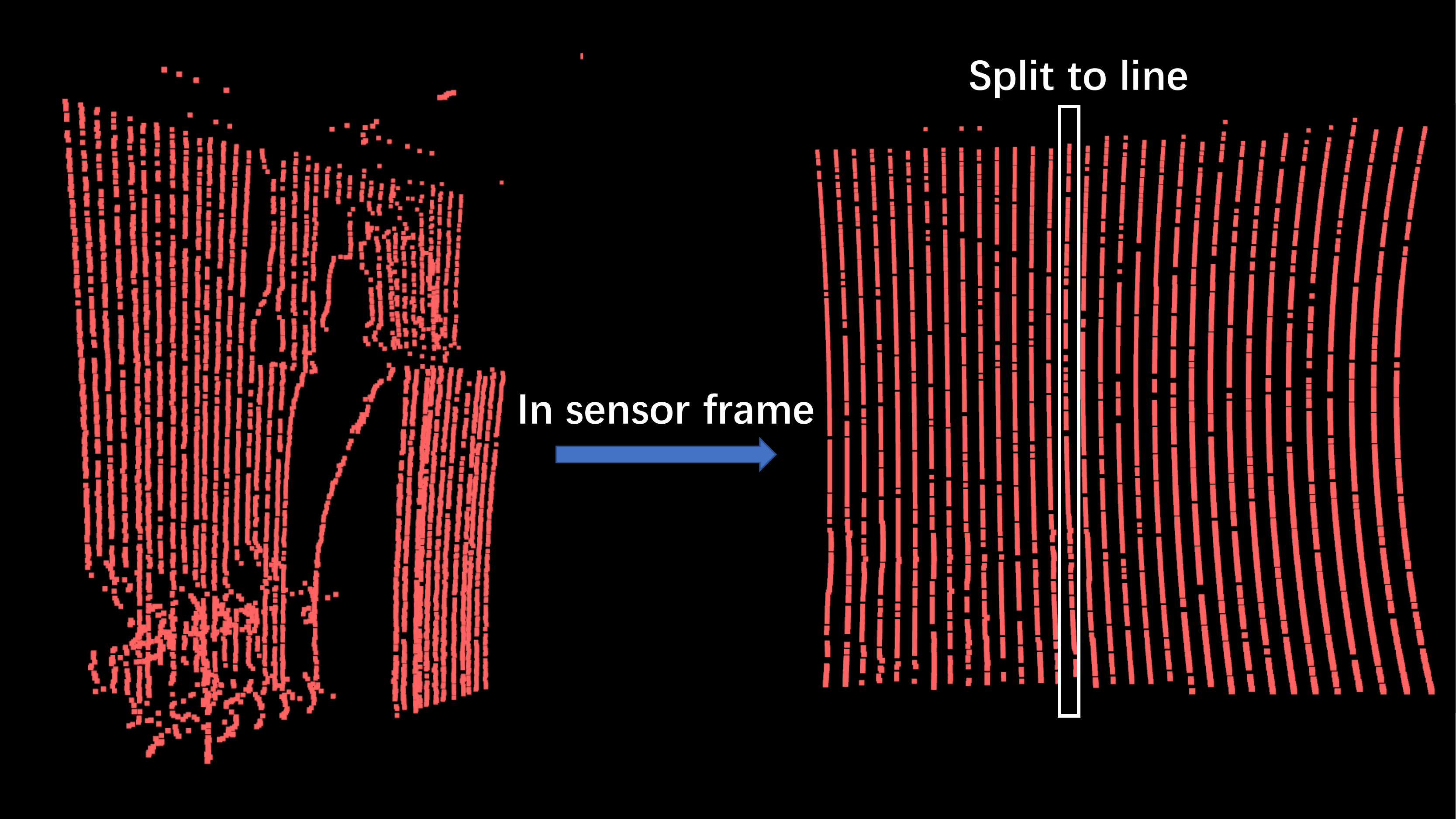}}\\
	\subfloat[Forward difference of one point line $l_i$. The red line is threshold. The bottom is on the left at index 0 and the top on the right.]{
		\label{fig:vertical_structure}
		\includegraphics[height=4.0cm]{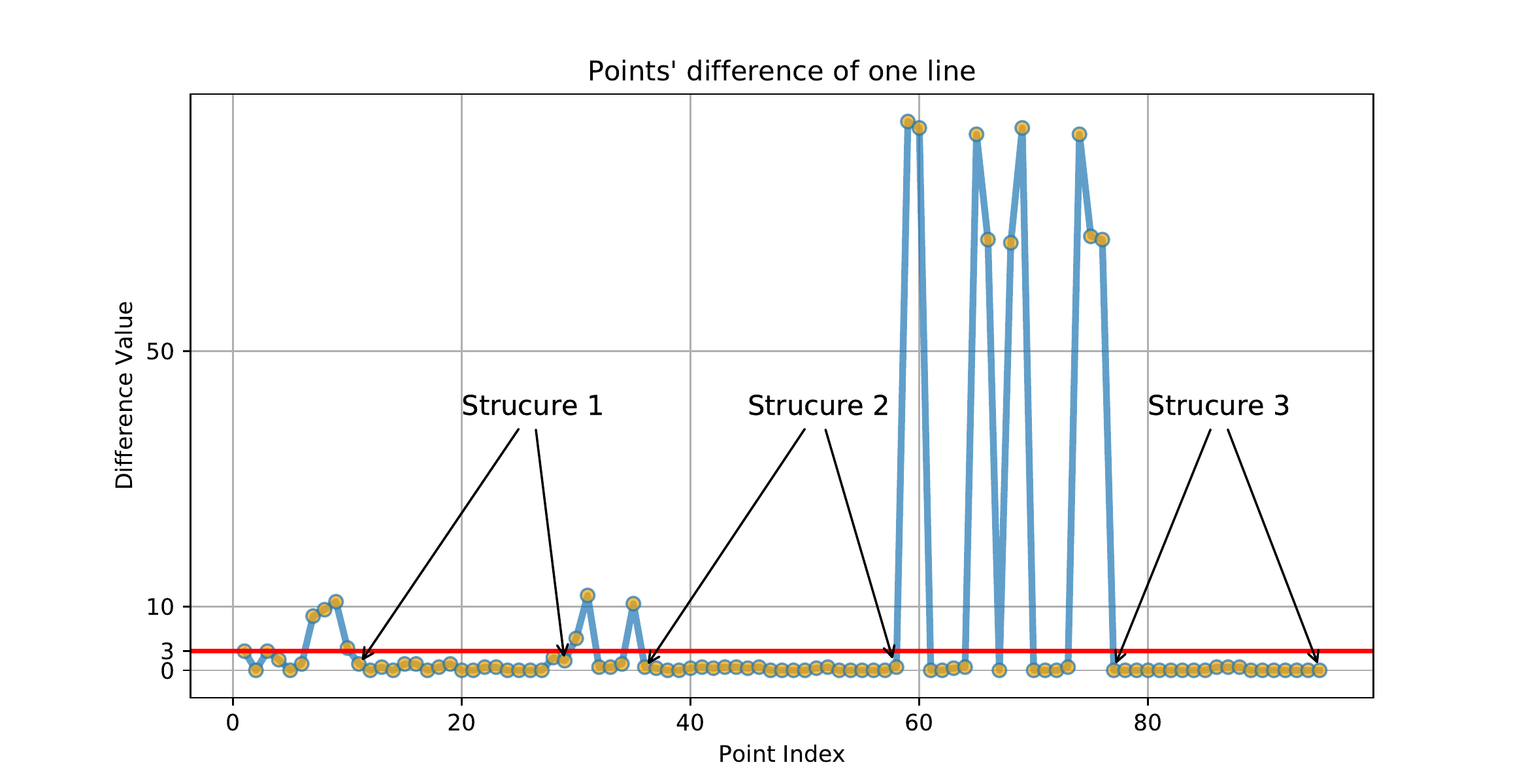}}
	\caption{Point line computation.}
\end{figure}

\begin{figure}[tb]
	\centering
	\subfloat[Up: The upmost planar structure of point lines in a scan. The 3D point cloud generated from several scans on the left and the 2D map on the right.]{
		\label{fig:up}
		\includegraphics[width=0.49\linewidth ,height=2.9cm]{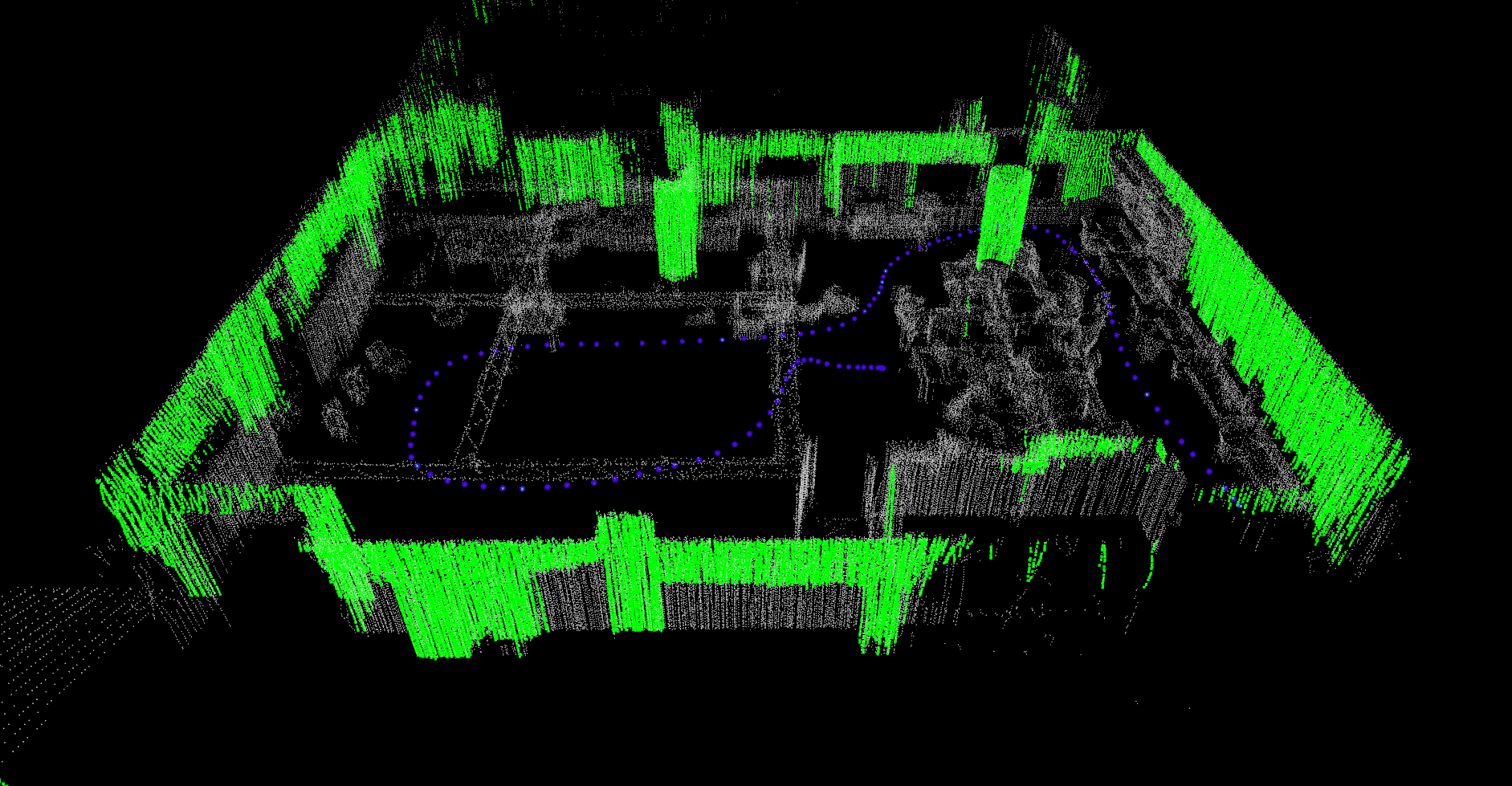}
		\includegraphics[width=0.49\linewidth ,height=2.9cm, angle=180, origin=c]{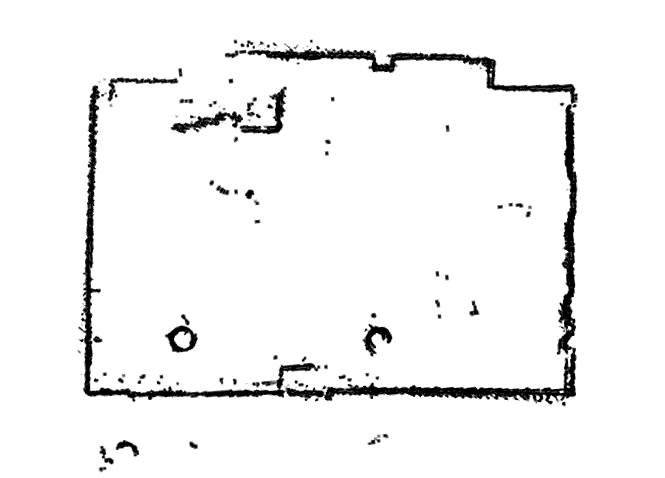}
	}\\
	\subfloat[Down: The bottom-most planar structure of point lines in a scan.]{
		\label{fig:down}
		\includegraphics[width=0.49\linewidth ,height=2.9cm]{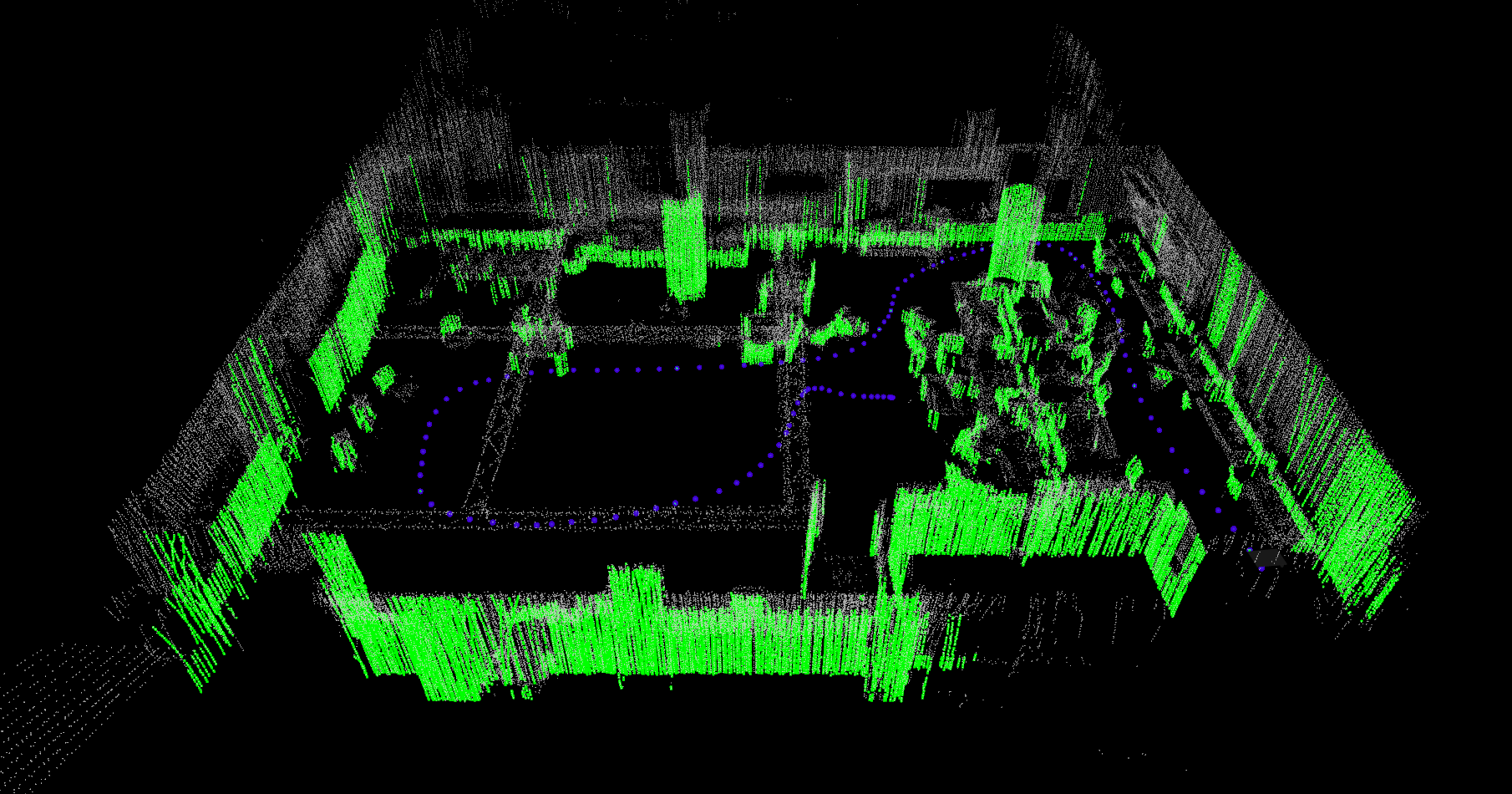}
		\includegraphics[width=0.49\linewidth ,height=2.9cm, angle=180, origin=c]{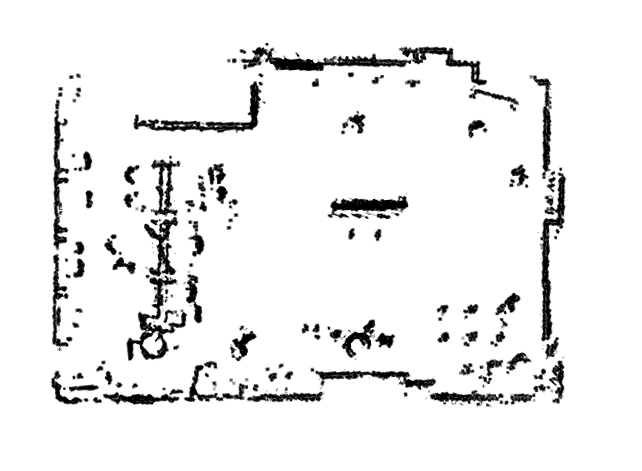}}	
	\caption{Planar structures in point lines and the corresponding 2D maps.}
	\label{fig:vertical}
\end{figure}

We show the forward difference of one point line $l_i$ in Fig. \ref{fig:vertical_structure}. It is the difference in the measured distance along the $Z$ axis. We apply a $threshold$ (shown as a red line) to separate planar structures of a certain minimum number of points (10 in these experiments). In the example, we can find three vertical structures in the line. Next, we need to choose the vertical structure that is most likely to be the wall. We assume that the highest detected structure is more likely to be the wall. This assumption is based on architectural design and space using practices: Typically, furniture does not go all the way up to the ceiling, so the top planar part should be the wall. If the furniture goes all the way to the ceiling, our algorithm will classify it as a wall. We think this is OK, because a cabinet that is that high is very unlikely to be moved. Fig. \ref{fig:up} shows the highest detected vertical structure in green and, for comparison, Fig. \ref{fig:down} highlights the bottom-most detected structure. We can see that Fig. \ref{fig:up} with the upper vertical structure shows a good representation of the wall. So we keep the upper detected vertical structure to be the wall candidate elements. 

\begin{figure}[htbp!]
	\centering
	\includegraphics[width=0.95\linewidth ]{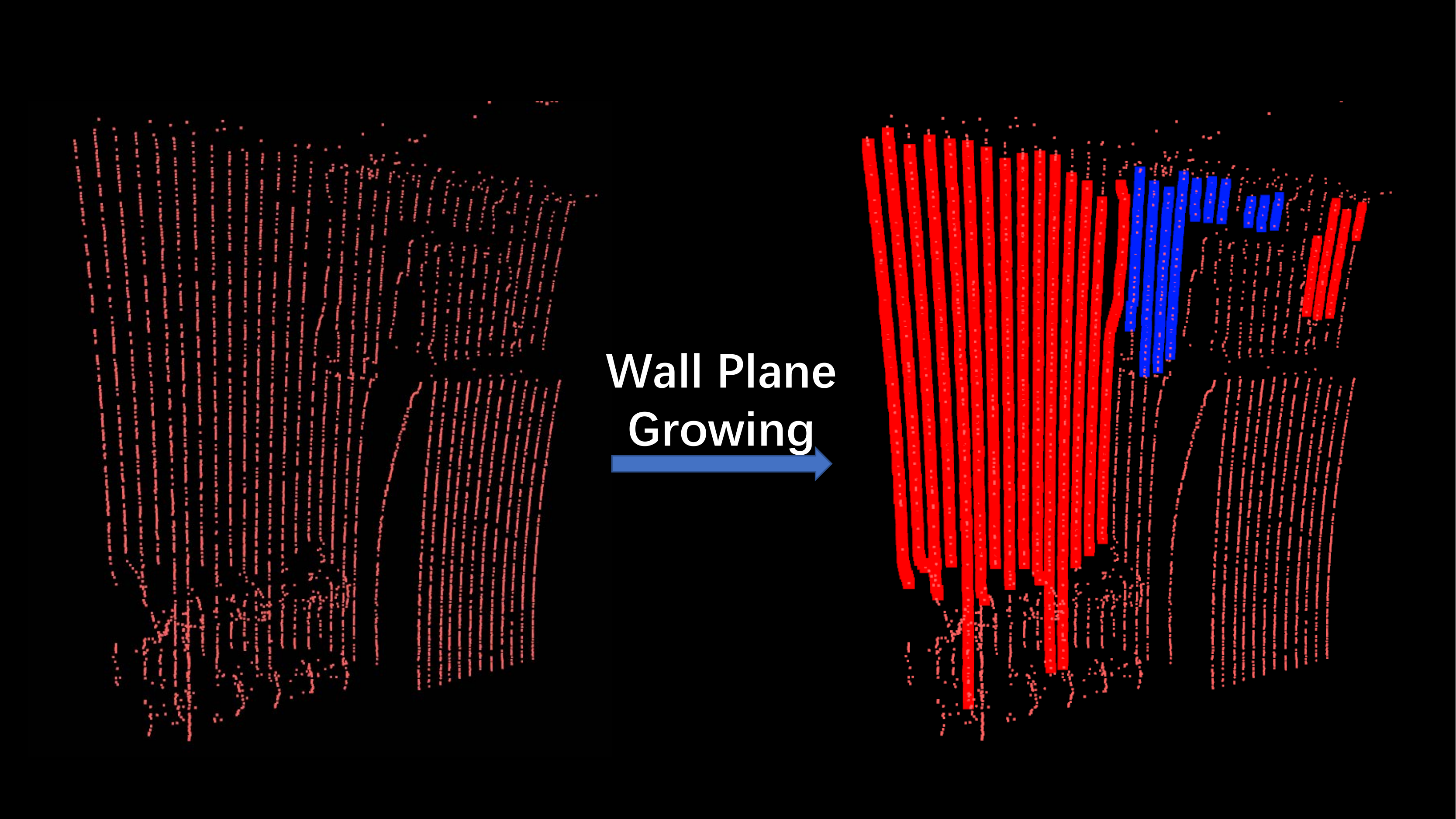}
	\caption{Find wall plane form the wall candidate elements.}
	\label{fig:origion}
\end{figure}

Afterwards, the growing method Alg. \ref{alg:WPD} is used to find the wall planes, which consist of many point lines. We pick a wall candidate as a seed and perform an iterative search for wall candidate elements that are coplanar with it, as shown in Fig. \ref{fig:origion}. In the end, we filter out the wall candidate elements which do not belong to the wall and calculate the wall plane parameters.

\begin{algorithm}[H]
	\caption{Wall Plane Detection}
	\label{alg:WPD}  
	\KwIn{Point lines set: $ \mathcal{L}$}
	\KwOut{The wall planes set $ \mathcal{W}$}
	\KwData{ \\ $ \mathcal{W} \leftarrow \emptyset $ \\
		$\mathcal{L} \leftarrow \{l_i, i \in \mathcal{N} \}$\\
		Current plane parameters $\mathcal{C} \leftarrow \emptyset$   \\
		Current plane $\mathcal{P} \leftarrow \emptyset$ \\
		Ransac function $\Omega(\cdot)$\\
		Similarity function $\Theta(\cdot,\cdot)$\\
		Similarity threshold $\sigma_{th}$   }
	\Begin{
		\For{i=1 to size($\mathcal{L}$)} {
			$l_i \leftarrow$ Wall candidate points $\rho_i$
		}
		\For{i=1 to size($\mathcal{L}$)} {
			\If{$\mathcal{P}$==$\emptyset$}{
				$\mathcal{P} \leftarrow l_i$
			}
			\ElseIf{$\mathcal{C}$ == $\emptyset$}{
				$\mathcal{C}$ == $\Omega(\{P\})$}
			\ElseIf{$\Theta(C,l_i) < \sigma_{th}$}{
				$\mathcal{P} \leftarrow l_i $ \\
				$\mathcal{C} = \Omega(\mathcal{P}) $
			}
			\Else{ $\mathcal{W}\leftarrow  \{\mathcal{P}\} \\
				\mathcal{P} \leftarrow \emptyset \\
				\mathcal{C} \leftarrow \emptyset $
			}
		}
		Return $\mathcal{W}$
	}
\end{algorithm}

\subsection{Semantic Labeling}

\begin{figure*}[htbp!]
	\centering
	\subfloat[]{
		\label{fig:3d_c}
		\includegraphics[width=0.19\linewidth ,height=3.5cm]{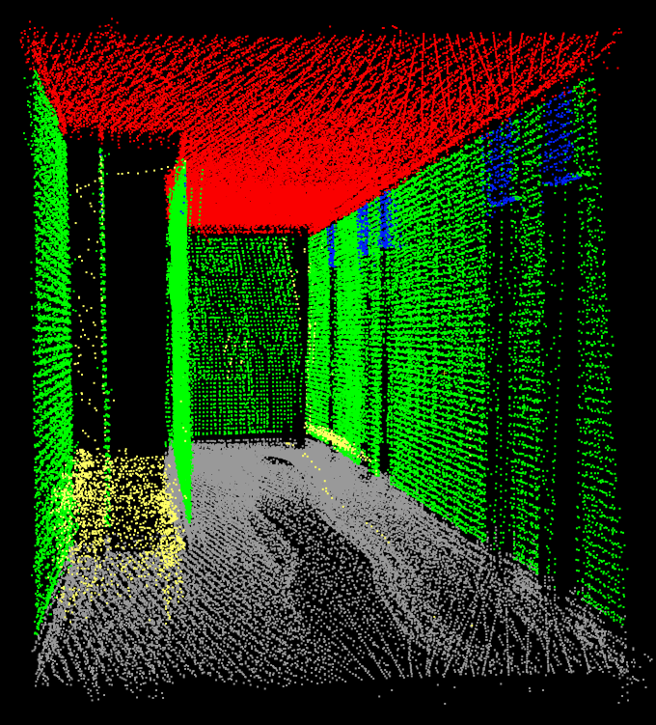}}
	\subfloat[]{
		\label{fig:3d_c_r}
		\includegraphics[width=0.19\linewidth ,height=3.5cm]{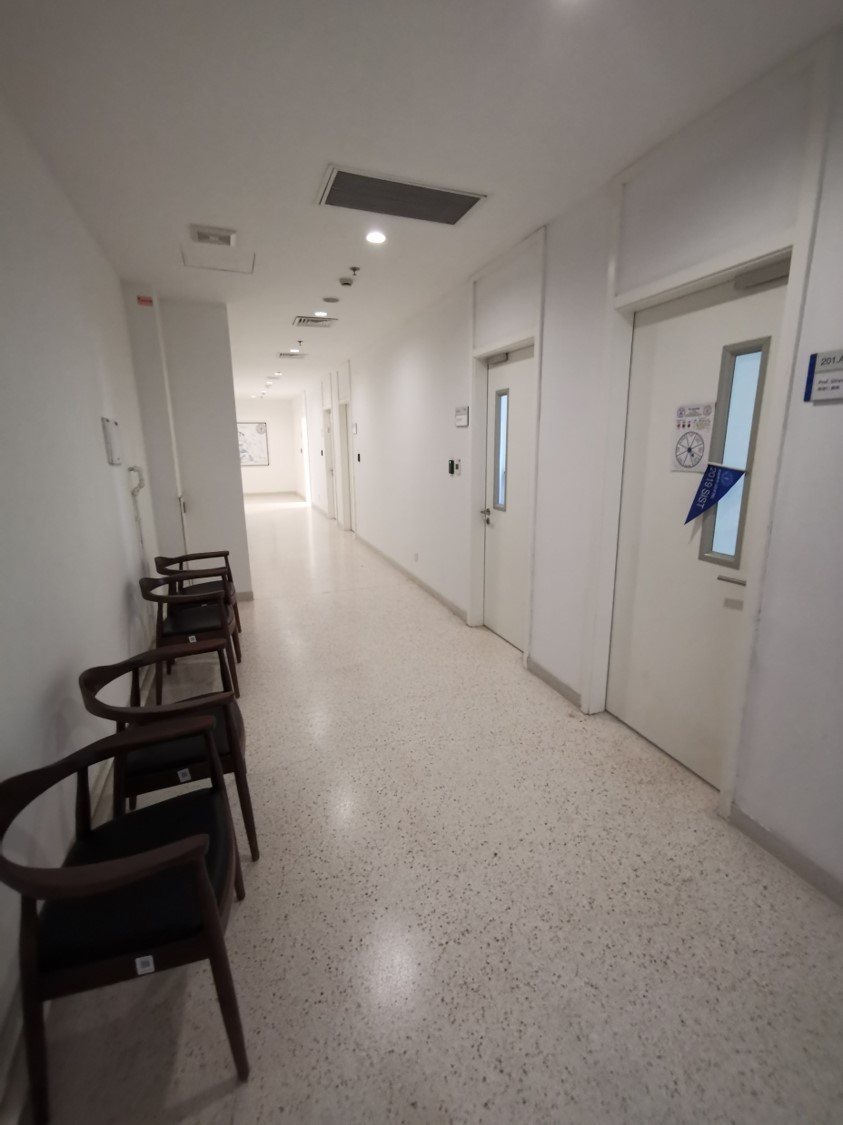}}
	\subfloat[]{
		\label{fig:3d_r}
		\includegraphics[width=0.3\linewidth ,height=3.5cm]{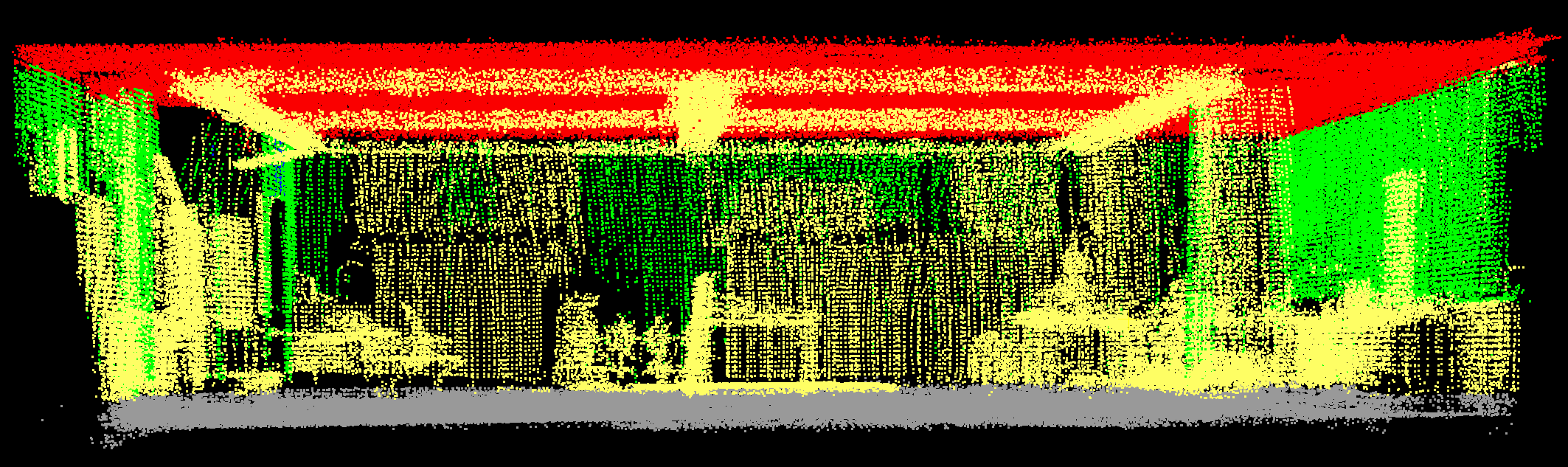}}	
	\subfloat[]{
		\label{fig:3d_r_r}
		\includegraphics[width=0.28\linewidth ,height=3.5cm]{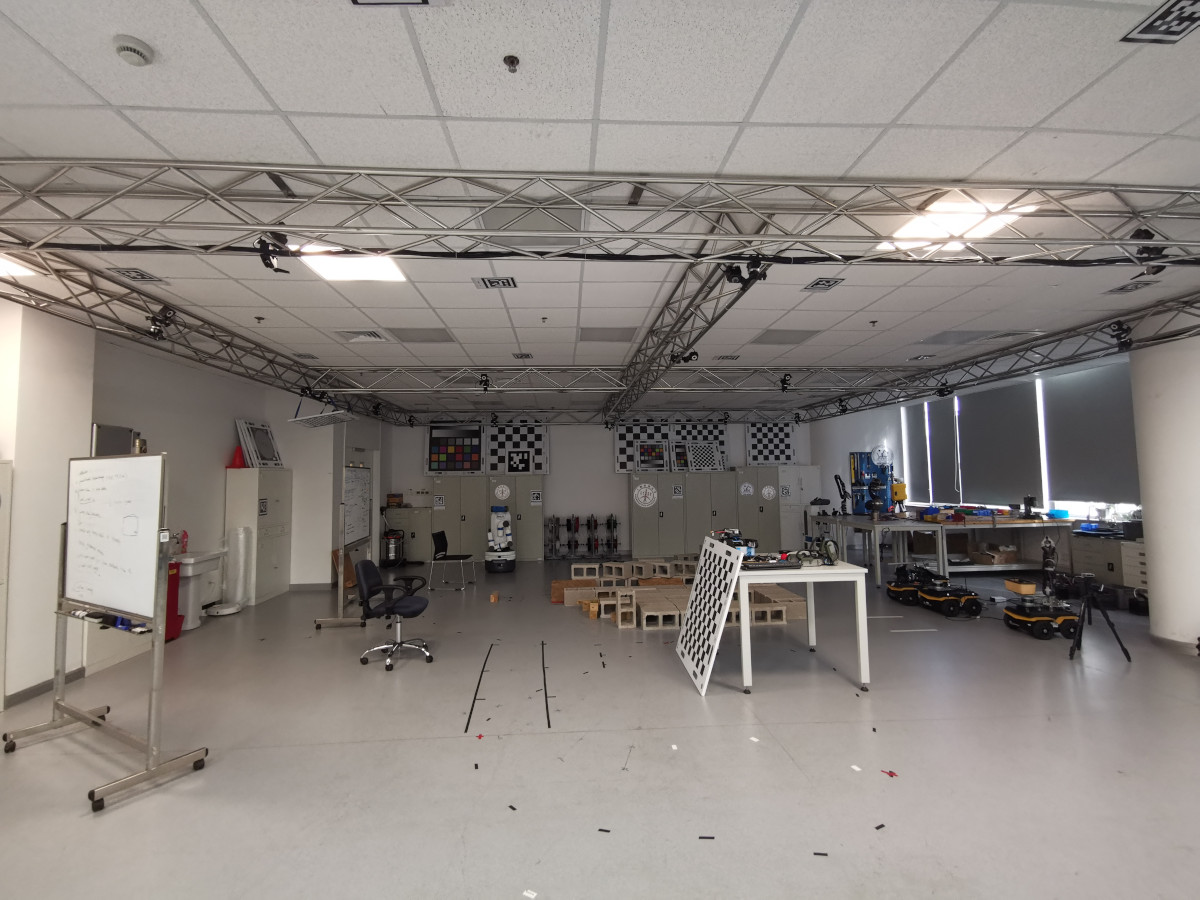}} \\ 
	\subfloat[]{
		\label{fig:3d_all}
		\includegraphics[width=0.75\linewidth ,height=4.25cm]{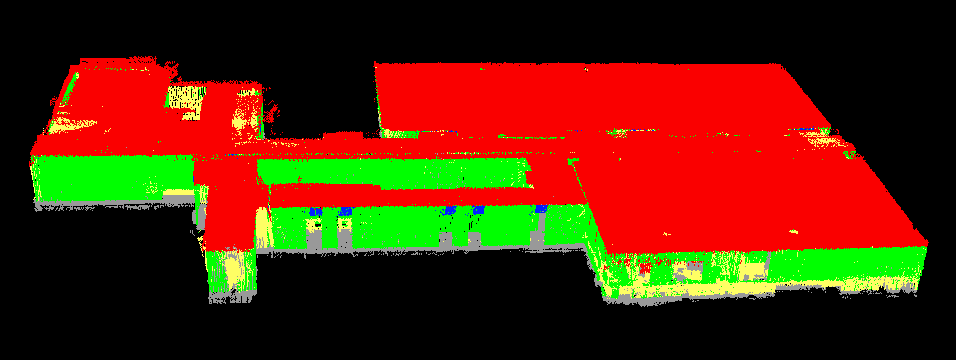}}
	
	\caption{(a) A corridor in our 3D result. (b) Real scene of the corridor. (c) Our robotics lab as 3D result. (d) Real scene of the lab. (e) 3D result of our process. All point clouds are the combined results from many individual, classified scans. Point color index: red: ceiling; gray: floor; green: wall; blue: door; yellow: other points.}
\end{figure*}

In the previous sections, we already labeled the ceiling, floor, and walls. The remaining points are labeled clutter. For robotics and mapping, doors are essential, as they are the passages between rooms, corridors, and other areas. Up to now, the wall segments above doors will just be classified as the wall. We are thus applying a dedicated door detection algorithm, that works for open and closed doors. 

The algorithm uses rule-based context knowledge to find the door. Usually, closed doors are a few centimeters deeper than the wall plane. Our algorithm can only detect this kind of closed door in the wall. Open doors have a very high hollow and are connected to the floor, whereas a closed door is just a depression on the wall. The algorithm works on the point lines from Section \ref{sec::rearrangement}. In order for that line to be classified as being above a door, all of the following conditions have to be met:

\begin{enumerate}
	\item   The points from that line belong to a wall plane.
	\item   Going from top to bottom, from some point onwards, all points of that line are located behind the plane of 1). 
	\item   This starting point from 2) has to be at least a minimum height above the ground. 
\end{enumerate}
Because our plane parameters are accurate enough, our method can still detect closed doors.

\section{Experiments and Discussion}

\subsection{Setup}

We evaluated our process in the robot with two Velodyne 32-line laser scanners. The robot mapping scenario includes several offices and corridors from the ShanghaiTech Automation and Robotics Center (STAR Center). Those are quite diverse spaces: cluttered offices, long and mostly empty hallways, a busy lab and a more empty lab. The dataset has closed and open doors. We cannot use publicly available datasets for comparison, as we rely on the vertical Lidar data, which other datasets don't have. Table \ref{tab:statistics} shows some statistics from the experiments. We show the  run time to compute the wall plane and door points of our single-threaded ROS C++ implementation on an Intel 3.60GHz 8 core laptop. We also show the number of points in the raw point cloud and the total number of points in the point lines (without ceiling and floor) per scan. 

\begin{table}[tb]  
	\centering  
	\begin{threeparttable}  
		\caption{Experiment Statistics}  
		\label{tab:statistics}  
		\begin{tabular}{ccccc}  
			\toprule  
			&average&std. err.&min&max\cr  
			\midrule  
			Run-time per cloud (ms) &74&26&32&187\cr   
			\#points in cloud&96,058&5,743&60,130&100,538\cr  
			\#points in lines&40,595&13,047&14,885&75,698\cr  
			\bottomrule  
		\end{tabular}  
	\end{threeparttable}  
\end{table}  

The labeling results for point clouds from this environment are shown in Fig. \ref{fig:3d_all}. These segmented point clouds are labeled into the ceiling (red), wall (green), door (blue), floor (gray), and clutter/ furniture (yellow). 
We are using the BLAM \cite{TPtoolboxweb}  SLAM algorithm with the horizontal Velodyne to provide the poses of the scans. With this information, we can then transform all labeled point clouds from the vertical scanner into a common coordinate system and thus create big point clouds, which we show in Fig. \ref{fig:3d_all}. By hand, we removed error points outside of the room, which are mainly a result of reflections of the laser beams on the windows. In the future we will do this automatically using this \cite{zhao2019mapping} approach. 

\subsection{Evaluation of 3D Point Cloud Semantic Labeling}
We use the recall, precision, and $F_{1-measure}$ value to quantitatively evaluate the correctness and accuracy of our 3D labeling result. The precision value describes the percentage of true objects in detected result, and the recall describes the percentage of true objects in all objects in the environment. The equations for recall, precision, and $F_{1-measure}$ are shown below:
\begin{equation}
\begin{aligned}
&&  &recall = \frac{TP}{TP+FN}  \\
&&\ &precision = \frac{TP}{TP+FP} \\
&&\ &F_{1-measure} = \frac{2*precision*recall}{precision+recall}
\end{aligned}
\end{equation}

Where FP, TP, FN are the number of false positives, true positives and false negatives, respectively. Because our algorithm is based on each frame of the point cloud, some data can be too noisy to detect the label. We use the detected area instead of the number of points to calculate these measurements. The quantitative evaluation is shown in Table \ref{tab:result}.  The recall and precision values of the wall are $95.26\%$ and $99.60\%$, respectively. There are only a few wrong labels in the 3D wall labeling result. However, our method cannot handle corners that are not fully scanned. There are also some problems with surfaces such as columns. For the ceiling, the recall and precision values are $85.25\%$ and $99.92\%$, respectively. The precision value is not high enough because we are not dealing with the points that are reflected by the window. In the future we will remove these outside points using \cite{zhao2019mapping}. With a height filter, we can get a high precision on the floor data. This approach cannot handle the uneven ground. The recall and precision values for door detection are $80.76\%$ and $82.82\%$, respectively. This is because some of the point lines are not vertical to the floor. In some situations those lines bend so that we mark the wall next to the door as a door. In this experiment there are 19 doors and we only missed one far away door in this scenario. So our approach works well to find out where the door is.

\subsection{Evaluation of Furniture Free 2D Maps}

\begin{figure}[tb!]
	\centering
	\subfloat[FARO ground trtuh.]{
		\label{fig:2d_gt}
		\includegraphics[width=0.47\linewidth]{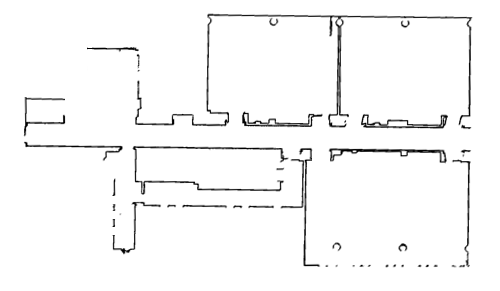}}
	\subfloat[Our result.]{
		\label{fig:2d_our}
		\includegraphics[width=0.47\linewidth]{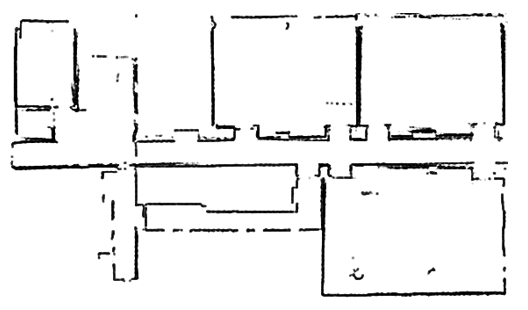}}\\	
	\subfloat[Cut below ceiling.]{
		\label{fig:2d_5060}
		\includegraphics[width=0.47\linewidth]{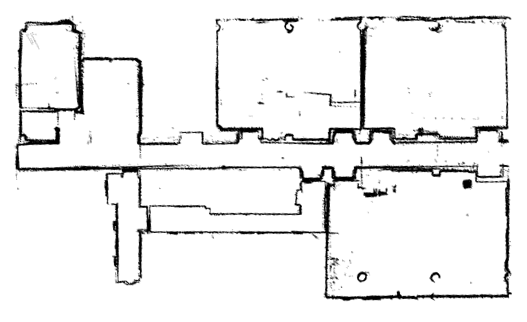}}
	\subfloat[Normal 2D mapping.]{
		\label{fig:2d_normal}
		\includegraphics[width=0.47\linewidth]{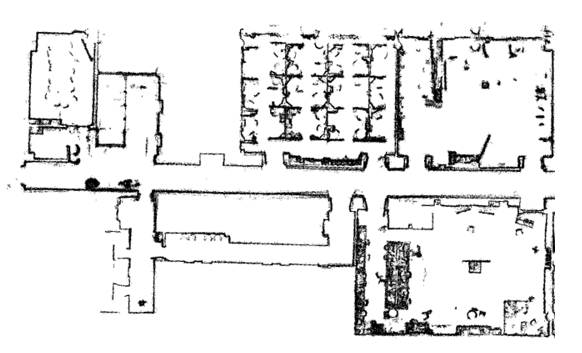}}\\
	\subfloat[Overlay of FARO and our map.]{
		\label{fig:2d_overlap}
		\includegraphics[width=0.47\linewidth]{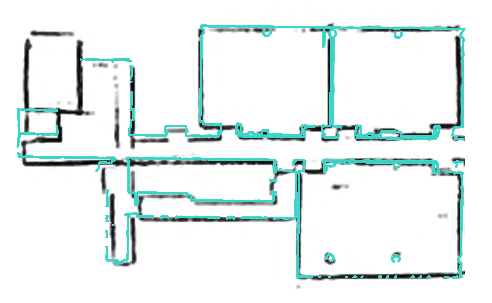}}
	
	\caption{Comparison of the different 2D grid maps of the STAR Center.}
	\label{fig:2d_compare}
\end{figure}

One main part of our result is the 2D furniture free map. We show several different 2D maps in Fig. \ref{fig:2d_compare}. Fig. \ref{fig:2d_gt} is a hand-made 2D map which comes from FARO. FARO is a high precision static 3D laser scanner with very low error. The FARO scans were taken when there was no furniture in the building, and all the doors were opened, so it is the perfect comparison. Fig. \ref{fig:2d_our} is our result. We observe that some doors are not open, even though they were classified as open in the 3D map. This error comes from the fact that doors should be detected as doors from both sides if there are scans at both sides. This is something to improve in the future. Fig. \ref{fig:2d_5060} is obtained by slicing 50 cm to 60 cm under the ceiling. In this map most of the furniture is removed, but there is still much noise in the environment. At the same time, the doors are not detected. Fig. \ref{fig:2d_normal}  is obtained by slicing the point cloud in the middle. It can keep the open doors open, but it is full of furniture, and closed doors are obstacles. In Fig. \ref{fig:2d_overlap} we intuitively compare the 2D furniture free map with the 2D map, which comes from FARO. The overlaid map shows that the room location is not fully aligned. This is because we have some drift in the SLAM process, and the poses from the SLAM Lidar are not from the same time as the scans of the vertical scanner. In the future we will interpolate the vertical Lidar pose from the surrounding horizontal Lidar poses. Our map accuracy is completely dependent on the accuracy of the SLAM process.

\renewcommand{\arraystretch}{1.5} 
\begin{table}[tb]  
	\centering  
	\fontsize{6.5}{8}\selectfont  
	\begin{threeparttable}  
		\caption{3D semantic labelling result of our method.}  
		\label{tab:result}  
		\begin{tabular}{ccccccc}  
			\toprule  
			\multirow{2}{*}{Label}&  
			\multicolumn{3}{c}{Label(area$/m^2$)}&\multicolumn{3}{c}{Result}\cr  
			\cmidrule(lr){2-4} \cmidrule(lr){5-7}  
			&FP&TP&FN&Precision&Recall&F1-Measure\cr  
			\midrule  
			Ceiling&141.68&819.15&0.60&85.25&99.92&92.00\cr   
			Floor&72.631&622.20&0.96&89.55&99.85&94.42\cr  
			Wall&2.80&703.24&34.99&99.60&95.26&97.37\cr  
			Door&3.77&18.17&4.33&82.82&80.76&81.77\cr  
			\bottomrule  
		\end{tabular}  
	\end{threeparttable}  
\end{table}

\begin{figure}[tb!]
	\centering
	\subfloat[]{
		\label{fig:area_normal}
		\includegraphics[width=0.49\linewidth ,height=2.95cm]{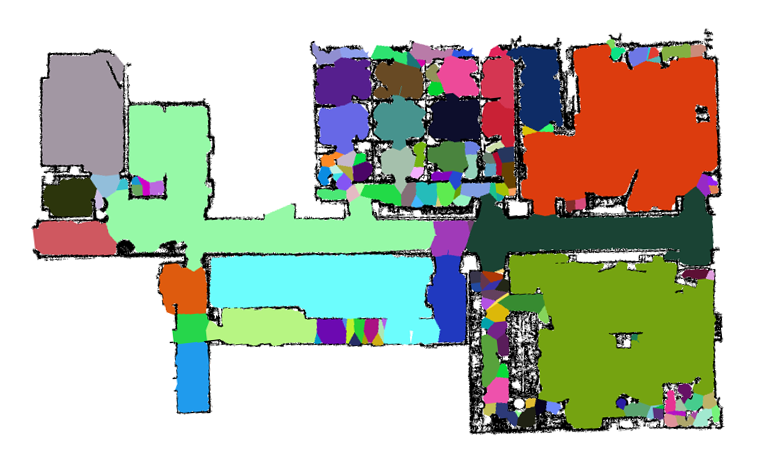}}
	\subfloat[]{
		\label{fig:area_our}
		\includegraphics[width=0.49\linewidth ,height=2.95cm]{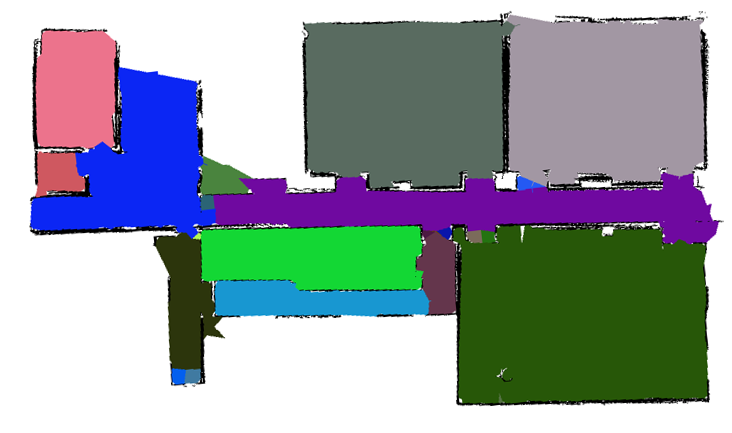}}
	\caption{Room segmentation applied on (a) the normal map with furniture and (b) the furniture free map, where each color presents an area.}
	\label{fig:areaGraph}
\end{figure}

\subsection{Application to Map Segmentation}

In \cite{hou2019area} we presented a topological map representation based on areas, the Area Graph. Figure \ref{fig:areaGraph} shows the Area Graphs generated from the normal map with furniture (Fig. \ref{fig:2d_normal}) and the furniture free map generated by our algorithm (Fig. \ref{fig:2d_our}). The parameter to segment the map is set according to the strategy described in \cite{hou2019area}  i.e., $ w = 1.9m$. We used an alpha shape removal approach to remove noise for the map with furniture, while we did not need use it for the furniture free map. It is straightforward to obverse that the Area Graph segmentation based on the normal map suffers from heavy over-segmentation. The segmentation on the furniture free map involves no over-segmentation of the rooms at all. We can also apply Area Graphs of furniture free maps in 2D map matching \cite{Hou2019Fast}.

\section{Conclusions}
This paper presented an efficient approach to automatically generate furniture free maps of indoor buildings at real-time speed. Removing the furniture in the rooms allows for simple tasks such as segmentation, map matching, and long-term localization. We achieve good labeling, also for doors, which is important for applications like path planning or topology generation. The method relies on a dense enough point cloud with coverage from the floor to the ceiling. Some limitations of our algorithm are: first, we need a good SLAM process to get localization, because the accuracy of the furniture free map depends on the accuracy of the slam algorithm. Second, there are usually lots of windows in buildings. The reflection of the laser beams window is hard to deal with. So we are also working on detecting and removing those erroneous beams. We also need to generate the model of rooms to optimize our result and to improve the door representation of furniture 2D grid maps, for example, by explicitly opening a door as soon as it is detected from one side.


\bibliographystyle{IEEEbib}
\bibliography{refs}

\end{document}